\documentclass[runningheads]{llncs}

\usepackage{microtype}

\usepackage{tabularx}
\usepackage{graphicx}
\usepackage{amsfonts}
\usepackage{amsmath}

\newcolumntype{L}{>{\raggedright\arraybackslash}X} 
\newcolumntype{C}{>{\centering\arraybackslash}X} 
\newcolumntype{I}{>{\itshape\centering\arraybackslash}X} 

\usepackage{graphicx}
\begin{document}

\title{Exploring Fluent Query Reformulations with Text-to-Text Transformers and Reinforcement Learning}
\titlerunning{QRT5}
%
\author{\thanks{Work done during internship at Vanguard \\ Workshop on the 9th Dialog System Technology
Challenge (DSTC-9), \\
Association for the Advancement of Artificial Intelligence 2021
(www.aaai.org)}Jerry Zikun Chen\inst{1,2}
\and Shi Yu\inst{1} \and
Haoran Wang\inst{1}}
%
%

\institute{Center for Analytics and Insights, The Vanguard Group, Valley Forge, U.S. \\
\email{\{jerry\_chen, shi\_yu, haoran\_wang\}@vanguard.com}\\
\and Department of Computer Science, University of Toronto, Toronto, Canada\\
\email{jzchen@cs.toronto.edu}
}

\maketitle              

\begin{abstract}
Query reformulation aims to alter noisy or ambiguous text sequences into coherent ones closer to natural language questions. This is to prevent errors from propagating in a client-facing pipeline and promote better communication with users. Besides, it is crucial to maintain performance in downstream environments like question answering when rephrased queries are given as input. We show that under the previous framework (AQA), attempts to alter RL algorithms do not bring significant benefits to either reward acquisition or sequence fluency. Instead, we leverage a query-reformulating text-to-text transformer (QRT5) and apply policy-based RL algorithms to further nudge this reformulator and obtain better answers downstream by generating reward-acquiring query trajectories. QRT5 shows better sample efficiency in RL to achieve the same level of QA performance as the previous approach. It can generate reformulations with more readability based on query well-formedness evaluations and can generalize to out-of-sample data. Our framework is demonstrated to be flexible, allowing reward signals to be sourced from different downstream environments such as intent classification.

\end{abstract}

\section{Introduction}
Query reformulation and paraphrase generation techniques are employed for a variety of purposes in natural language processing (NLP), such as dialogue generation \cite{Liu2018QuestionRB}, machine translation \cite{Madnani2010TheCO}, and especially in question answering (QA) systems \cite{Figueroa2013LearningTR,Witteveen_2019,Elgohary2019CanYU}. Generating coherent and clean texts reduces potential errors in downstream systems. In the case when users are at the receiving end of NLP pipelines, it is essential to show fluent and human-like languages before trust is lost until a point where users recede into requiring human agents for the sake of better communication. Lastly, by having a reformulator model to modify queries, results can be fed back to users and confirm their original intentions in an automated way.

The advent of Seq2Seq learning \cite{Sutskever2014SequenceTS} made it possible to train deep neural networks as a new paradigm to replace rule-based and statistical approaches to generate reformulations and paraphrases \cite{Meteer1988StrategiesFE,McKeown1979ParaphrasingUG,Zhao2009ApplicationdrivenSP}. We investigate how to generate well-formed reformulations using Seq2Seq models that can maintain good QA performance at the same time. We apply the pre-training and RL pipeline from previous work AQA \cite{Buck2018AskTR} to the T5 framework \cite{Raffel2019ExploringTL} and fine-tune the QRT5 reformulator on paraphrasing and denoising, before RL is applied in downstream QA and intent classification (IC) environments that provide reward signals. We choose T5 because it is a state-of-the-art Seq2Seq model that suits the query reformulation task. It provides flexibility for tuning on different datasets without changing the training pipeline significantly. Furthermore, the self-supervised pre-training \cite{Erhan2010WhyDU} and sequential transfer learning approach \cite{Ruder2019Neural} in T5 has consistently provided good inductive bias in the past, obtaining improvements across many NLP benchmarks \cite{Devlin2019BERTPO,liu2019roberta,Raffel2019ExploringTL} and showing great out-of-distribution generalization \cite{hendrycks2020pretrained,brown2020language}. 

To our knowledge, this is the first attempt at training T5 with RL to produce query reformulations. We show that QRT5 is a better starting point for RL and more sample efficient to achieve the same level of QA performance as AQA. The efficiency of RL is important in a productionized pipeline where rewards are defined by humans. These systems in practice are typically expensive to interact with when they are black boxes that can be frequently updated or changed. This is even more relevant to query reformulation as episodic rewards from QA or IC are produced only after a complete sequence is constructed, without intermediate signals in between tokens to updates the model. In addition, QRT5 reformulates with better readability and can generalize to out-of-distribution (OOD) data. This is crucial because queries have word token permutations and varying levels of ambiguity and syntactic complexity \cite{Mothe2005LinguisticFT} with distinct properties of their own \cite{Barr2008TheLS,Roy2013AnalyzingLS}. Thus, semantic meanings may be lost in reformulations and need to be preserved, especially for OOD queries while improving fluency. Lastly, we evaluate reformulation fluency using scores produced by a separate T5 model trained on the question well-formedness (QW) \cite{Faruqui2018IdentifyingWN} dataset, which is based on real evaluations from humans. Widely-used algorithmic heuristics based on overlapping n-grams like BLEU \cite{Papineni2002BleuAM} or ROGUE \cite{Lin2004ROUGEAP} are found to be less correlated with human judgements \cite{CallisonBurch2006ReevaluationTR,Stiennon2020LearningTS}. Query well-formedness training provides a proxy to sequence-level fluency that mimics human preferences.

\section{Preliminary}
\label{sec:training}
\subsection{Supervised Fine-tuning}
Given a query sequence $\mathbf{q}= \{q_1, \cdots, q_k\}$ of length $k$ in a dataset of size $N$, the reformulator model produces a sequence of word token distributions. A reformulation $\mathbf{r} = \{r_1, \cdots, r_T \}$ is produced by sampling tokens from these distributions at each time step. They are matched with the target sequence $\mathbf{y} = \{y_1, \cdots, y_T \}$ in a final loss. We assume here that $\mathbf{y}$ and $\mathbf{r}$ have the same length. Practically, sequences are padded to a default max length of 50, and any token distributions produced after the end special token are disregarded in the loss. Both of input $\mathbf{q}$ and target $\mathbf{y}$ are tokenized by a pre-defined sentence-piece \cite{Kudo2018SentencePieceAS} with vocabulary size $V$. We use the default vocabulary of size 16,000 for AQA and 32,168 for QRT5. For the $i$'th data point, the conditional sequence probability of the reformulation $\mathbf{r}^i$ is given by
$ \pi_{\theta}(\mathbf{r}^i | \mathbf{q}^i) = \prod_{t=1}^{T} p(r^i_t | r^i_1, \cdots, r^i_{t-1}, q^i_1, \cdots, q^i_k)$. The correct label $\mathbf{y}^i$ is defined by a sequence of one-hot encodings at each time step. Therefore, the cross entropy loss between the model predictions $\mathbf{r}^i$ and target $\mathbf{y}^i$ is:
$$\mathcal{L}_{CE} = - \sum_{i=1}^{N} \mathbf{y}^i \log \pi_{\theta}(\mathbf{r}^i | \mathbf{q}^i)$$
$$ =  - \sum_{i=1}^{N} \sum_{t=1}^{T} \sum_{j=1}^{V} y_{j, t}^{i}\log p(r_{j, t}^i | r_{1}^i, \cdots, r_{t-1}^i, \mathbf{q}^i)
$$
where $y_{j, t}^i \in \{0, 1\}$ is the binary label of token $j$ at time $t$, and $p(r_{j, t}^i | \cdot)$ is the conditional probability of token $j$ appearing at $t$ given previously produced tokens and input query. Note that minimizing cross entropy loss $\mathcal{L}_{CE}$ is equivalent to minimizing the negative log-likelihood $\mathcal{L}_{NLL} = - \sum_{i=1}^{N} \log \pi_{\theta}(\mathbf{r}^i | \mathbf{q}^i) = - \sum_{i=1}^{N} \sum_{t=1}^{T}  \log p(r_t^i | r_1^i, \cdots, r_{t-1}^i, \mathbf{q}^i) $

\subsection{Reinforcement Learning}
In the RL stage, we follow \cite{Buck2018AskTR} and optimize for the expected long term rewards: 
$$\mathcal{J} = \sum_{i=1}^{N} \mathbb{E}_{r^{i}\sim\pi_{\theta}} (\sum_{t=1}^{T} R(r^i_1, \cdots, r^i_t))$$
where $R$ is the black-box function that generates rewards between 0 and 1 only at the end of generation $t = T$. In the QA setting, this reward is the character-level F1 score $R_{F1} = 2 (p\cdot r) / (p + r)$ between the true answer $a$ and the answer output $a'$ from the pre-trained BiDAF QA model \cite{Seo2017BidirectionalAF}, when the reformulation is given as input. Precision $p$ is the proportion of tokens in $a'$ that are in $a$, while recall $r$ is the proportion of tokens in $a$ that are in $a'$. The main quantitative metric of interest is the F1 score on the dev set as this shows how well models can generalize on unseen SearchQA \cite{Dunn2017SearchQAAN} data during RL. 
From a batch of size $b$, the gradient of $\mathcal{J}$ can be estimated by REINFORCE \cite{Williams1991FunctionOU}, by sampling a reformulation trajectory $\mathbf{r}^i = \{r_{1}^i, \cdots, r_{T}^i\} \sim \pi_\theta$ given $\mathbf{q}^i$:
$$\nabla \mathcal{J} \approx \sum_{i=1}^{b} \nabla_{\theta} \log p(r_{1}^i, \cdots, r_{T}^i | \mathbf{q}^i)(R(\mathbf{r}^i)-B^i)$$
$$= \sum_{i=1}^{b} \nabla_{\theta} \mathbf{r}^i \log \pi_{\theta}(\mathbf{r}^i | \mathbf{q}^i)(R(\mathbf{r}^{i})-B^i)$$
This means that we can maximize the above weighted log likelihood function (i.e. minimizing a weighted cross-entropy loss $\mathcal{L}_{CE}$) as a surrogate to compute and estimate the policy gradient \cite{Sutton1999PolicyGM} in each batch. The target sequences in this surrogate cross entropy loss are the sampled reformulation trajectories $\mathbf{r}^i$ from the policy for estimating the expectation. A trajectory that obtains a higher reward will produce a higher gradient signal to encourage generation of words close to this sampled reformulation at each time step. Similar to AQA, for variance reduction, the mean reward of the minibatch is used as the baseline $B$ in the gradient and the loss. In addition, a scaled entropy regularizer $\lambda H(\pi_{\theta}) = \lambda \sum_{t}\sum_{j} p(r_{j, t} | \mathbf{r}_{<t}, \mathbf{q}^i) \log p(r_{j, t} | \mathbf{r}_{<t}, \mathbf{q}^i)$ is added to the loss to mitigate deterministic policy updates. The modifications in following sections originate from this objective function, which we refer to as the policy gradient (PG) baseline.

\section{Methodology}
\subsection{Datasets}
\label{sec:dataset}
\textbf{SearchQA} \cite{Dunn2017SearchQAAN} is comprised of 140,000 question-answer pairs from the Jeopardy! archive. Questions are inputs to the reformulators. In RL, the QA environment model has been pre-trained with this dataset on machine comprehension to extract answers within the context given a question. The queries and context snippets are mostly convoluted or ambiguous. \textbf{Paralex and UN Parallel Corpus} \cite{Fader2013ParaphraseDrivenLF,Ziemski2016TheUN} are paraphrasing and multilingual translation datasets used by \cite{Buck2018AskTR} for supervised pre-training. \textbf{Question well-formedness (QW)} dataset \cite{Faruqui2018IdentifyingWN} are filtered from Paralex, which we use for fine-tuning the well-formedness T5 model. Every query in QW is scored either 0 or 1 by 5 human workers. The average score is reported as the well-formedness ratings.
\textbf{Quora and MQR} datasets are used to fine-tune QRT5 from the pre-trained model. The Quora dataset contains similar yet differently expressed question pairs from the online Q\&A forum. The multi-domain question rewriting (MQR) \cite{Chu2020HowTA} dataset consists of ill-formed and well-formed question pairs, for example, \textit{``Spaghetti carbonara, mixing"} is paired with \textit{``How to mix a spaghetti carbonara?"}.
\textbf{An internal log dataset} of 300k queries with intent labels produced by human agents is leveraged for intent classification and out-of-sample analysis.

\subsection{AQA Framework}
\label{sec:aqa}
We directly use the same downloaded pre-trained checkpoint in \cite{Buck2018AskTR} as the starting model for different RL variants. Its architecture is based on GNMT \cite{Wu2016GooglesNM}, pre-trained on UN parallel corpus and Paralex for general paraphrasing capabilities. Note that we do not use the CNN selector from AQA as the focus is to produce a single reformulator. The same BiDAF QA model pre-trained on SearchQA is used as the black-box QA environment. The reformulator policy passes a reformulation batch to the QA environment to obtain a batch of character-level F1 scores. As mentioned, the baseline in this approach is the mean reward of the batch. In Section \ref{qualitative_comparison_sqa}, a downloaded RL-tuned AQA reformulator is also used to compare generation quality with other methods.

\subsubsection{Reinforcement Learning}
\label{sec:method_rl}
The advantage actor-critic approach is a common modification for policy gradient methods where $A^i = R(\mathbf{r}^i)-B^i$ is the advantage estimate. We use a two-layer neural net $f_c$ as an on-policy critic network to estimate the value for each batch. The output embedding from the GNMT encoder $\pi_\theta^E$ is used as the input for this critic network. The value becomes the baseline $B^i = f_c(\pi_\theta^E(\cdot | \mathbf{q}^i))$ and it is an estimation of the expected reward for the batch, given the current model's encoding of the input. 


We test another common approach to tune Seq2Seq models, the self-critical training \cite{Rane2018SelfCriticalST}. First proposed for image captioning, this method uses the reward generated by the greedy output $\mathbf{r}^i_{greedy} \sim \pi_{\theta}(\cdot | \mathbf{q}^i)$ as the baseline $B^i=R(\mathbf{r}^i_{greedy})$ in the policy gradient formulation. This is to encourage the model to outperform the greedy decoding strategy.


Beside varying the RL algorithms, we test methods that explicitly encourage fluency as reformulations produced by AQA often contain repetition of words and phrases. Unlikelihood training \cite{Welleck2020NeuralTG} is proposed as an extra term to regularize the loss function and explicitly suppress the likelihood of negative candidate tokens (repetitions) $\mathcal{C}^t = \{r_1, \cdots, r_{t-1}\} \backslash \{r_t\}$ in a reformulation sequence $\mathbf{r} = \{r_1, \cdots, r_T\}$. In this method, the following loss is weighted by the advantage estimate with the mean reward as the baseline:
$$\sum_{t=1}^{T} [- \alpha \sum_{c\in \mathcal{C}^t} \log(1 - p(c|\mathbf{r}_{<t}) + \mathcal{L}_{NLL}] \text{ ,}$$
$$\text{where }\mathcal{L}_{NLL}= - \log p(r_t|\mathbf{r}_{<t}, \mathbf{q})$$
Another addition is the fluency metric from \cite{Ge2018FluencyBL}, which is proposed for error correcting sequence generation and inference:
$$R_f = \frac{1}{1 + H(\mathbf{r})} \text{ ,} $$
where $H(\mathbf{r}) = -\frac{1}{T}\sum_{t=1}^T \log p(r_t | \mathbf{r}_{<t}, \mathbf{q})$. This metric ranges between 0 and 1 and incorporates the probabilities produced by the model. We use it as an extra reward signal on top of the F1 reward $R(\mathbf{r})$ from the QA environment.

\subsection{QRT5 Framework}
\label{sec:sft_t5}
With recent advances of transfer learning in NLP, we want to leverage general language representation power encoded in a pre-trained transformer-based model. Due to pre-training and architectural limitations of AQA, we leverage the T5 baseline (T5-base) model to replace the LSTM-based reformulators. Hugging Face \cite{Wolf2019HuggingFacesTS} and PyTorch Lightning \cite{falcon2019pytorch} are leverged in the implementation. This allows us to fine-tune on supervised tasks with more flexibility and to create our own starting points for RL. T5 has a similar encoder-decoder structure as the original transformer \cite{Vaswani2017AttentionIA}, which was designed for Seq2Seq tasks. T5 formulates any language tasks into the text-to-text format, where a prefix description of the task is attached to each input, instructing the model to perform the task through text without having to vary much of the training pipeline.

\subsubsection{Supervised Paraphrasing and Denoising}
Pre-training of Seq2Seq models before RL is a necessary step \cite{Buck2018AskTR,Jaques2017TuningRN,Choshen2020OnTW}. We tune a new RL starting point to replace the translation-based model from AQA that has 210 million parameters. This capacity is close to the T5-base model with 220 million parameters. It is suggested in \cite{Chu2020HowTA} that Quora (paraphrase) and MQR (denoising) create a good combination for improving query qualities so they are are used for fine-tuning. For both datasets, we add prefix \textit{``paraphrase: "} to the input sequence and special end suffix ``\textless$\backslash s$\textgreater" to both the source and target sequences. We lightly tune for two epochs respectively.

\subsubsection{Reinforcement Learning}
We adopt two RL approaches similar to those in the AQA framework mentioned before. In particular, since the policy gradient (PG) baseline shows the best QA performance in Section \ref{sec:rl_aqa_results}, we focus on this method for further RL of QRT5. Self-critical (SC) training, being another common approach with decent performance, is used as an alternative. For modularity, the implementation leverages the T5 module from Hugging Face. QRT5 continues the reformulation task during RL, so the same prefix and suffix are added as mentioned in the above section.

\subsection{T5 for Query Well-formedness}
\label{sec:wf}
For a non-algorithmic numeric proxy for sequence fluency, we use the QW dataset to fine-tune a separate T5-base model, which produces automatic scores. We cast regression as a text-to-text task by generating a text string of the average rating score and compare it against the true label, e.g., a score of 3.0 becomes \textit{``3.0"}. The prefix for this task is \textit{``query wellformedness: "}. Since the labeled scores are averaged among 5 humans, there are 6 categories from 0.0 to 1.0. Therefore, this is regarded as a 6-way supervised classification task in the text-to-text framework similar to how the STS-B regression task is formulated by \cite{Raffel2019ExploringTL}. This model judges how well-formed or fluent a given query is based on human evaluations. It is fine-tuned for 50 epochs on QW, when validation set accuracy no longer improves.

\subsection{Intent Classification}
To demonstrate the flexibility of QRT5, we experiment with a pre-trained BERT-based intent classification (IC) model \cite{yu2020financial} as another fixed downstream system to produce reward signals. We first use the fine-tuned QRT5 to reformulate every test set query naïvely in the internal query log dataset and measure accuracy and F1 scores on the intent classes predicted by the pre-trained IC model. We compare this to the original IC test set performance. In addition, RL is applied for 5 epochs on the training set to further adapt QRT5 to the IC enviornment. Rewards are engineered as follows: when the predicted intent class exactly matches the true class, QRT5 receives a reward of 1. Otherwise, leveraging the hierarchical structure of intent classes, if a match between the parent label of the predicted class and true class occurs, QRT5 gets a partial reward of 0.5. If none of the above, the IC system gives 0 reward. Again, we measure accuracy and F1 scores on the test set with reformulations produced by QRT5 with RL. These comparisons are meant to show whether or not QRT5's reformulations from interactive RL can be adapted to an black-box IC model that was pre-trained on the original noisy text, similar to the setup of QA. 


\section{Experimental Results}
\subsection{AQA Framework for RL}
\label{sec:rl_aqa_results}
\begin{figure}[!ht]
\begin{center}
\centerline{\includegraphics[width=80mm]{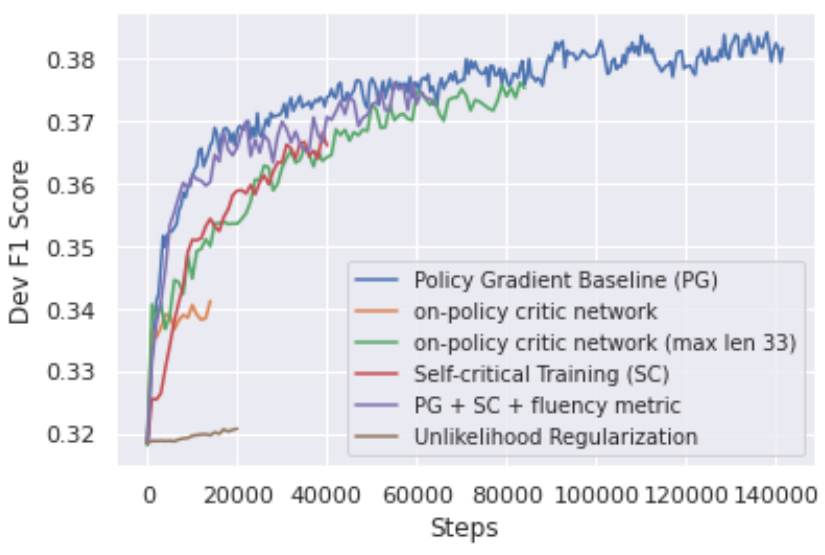}}
\caption{SearchQA Dev Set F1 Reward Curves with AQA}
\label{aqa_rewards}
\end{center}
\end{figure}

Figure \ref{aqa_rewards} plots the mean validation (dev) set F1 scores on SearchQA during RL of the reformulator. The baseline method is the original AQA approach and the rest are variations that we mentioned in the previous section. Most of the variations are able to adapt to the RL objective and learn from character-level F1 scores through trial and error except for the unlikelihood objective. It did not integrate with the model as well as we expected to reduce the number of repetitions and gain rewards. The addition of a critic network only surpasses the baseline slightly at first and becomes flat quickly. When we reduce the max length of the reformulation output produced by the translation model, the method with a critic network improves its performance. However, forcing the model to produce shorter reformulations is not ideal. The model must learn when to stop by itself. After RL, we notice most of these methods tend to generate long sequences close to the default maximum of 50, including the best-performing PG baseline. Furthermore, we observe variations in the RL algorithm fail to outperform the original PG baseline meaningfully in learning the reward under the AQA framework. The closest reward curve to the baseline method is having a mixed loss function that combines PG and SC objectives, with the addition of the fluency metric as an extra reward. In the next section, we experiment with QRT5 that can both retain fluency and optimize QA performance with RL. Compared to other methods except for the PG baseline, self-critical training, when combined with the policy gradient objective, performs the best under the AQA framework, and achieves the closest performance to the baseline. We adopt SC as an alternative to the PG method in the QRT5 framework.

\subsection{QRT5 Framework for RL}
\begin{figure}[!h]
\begin{center}
\centerline{\includegraphics[width=80mm]{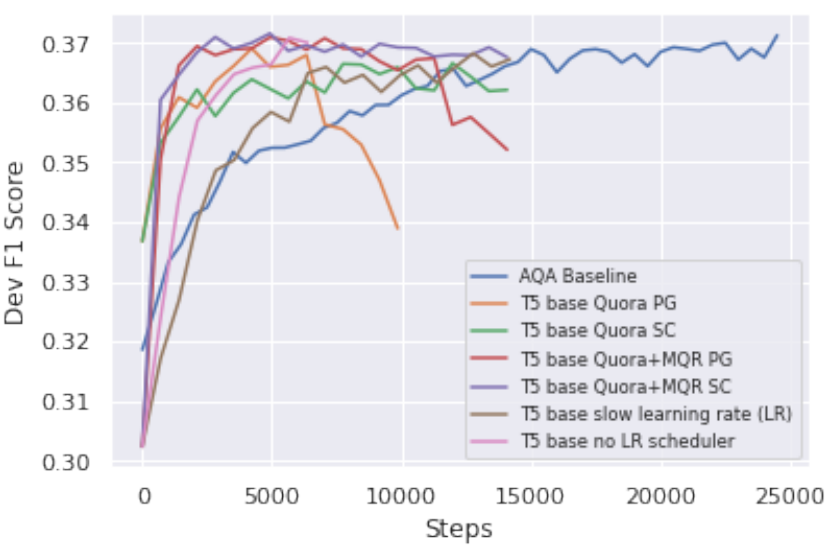}}
\caption{SearchQA Dev Set F1 Reward Curves with QRT5}
\label{t5_rewards}
\end{center}
\end{figure}
In Figure \ref{t5_rewards}, when QRT5 models are tuned with RL, we observe better sample efficiency with faster reward acquisition. QRT5's reward on the validation set is initially lower than the AQA baseline model. As RL progresses, the reward grows quickly after a few epochs. The reward curves for PG methods do eventually drop, and the SC methods turn flat. When the gradient is tracked, its norm becomes large. We suppose this is due to RL optimization taking steps in adverse directions and not able to recover. Note that we track the reward curves on the validation set rather than the training set to evaluate generalization. The training set reward curves do not drop, which suggests that overfitting can occur in the RL stage for T5 models. When the reward drops on the dev set, we observe the model produces more repetitions similar to AQA, losing query fluency. Changing the learning rate scheduler, entropy hyperparameters, or learning rates do not resolve this issue. Therefore, we pick the best-performing PG model with dev set F1 score around 0.37 as our model for qualitative analysis in Section \ref{qualitative_comparison}. Although the T5 model cannot train for longer without sacrificing rewards, it learns almost 3 times faster than the AQA model and reaches a comparable QA performance with only 0.01 difference in F1 as AQA reformulator eventually reaches 0.38 dev F1 score after 9 days of training on a single Tesla T4 GPU.

\subsection{Well-formedness}

We fine-tune a T5-base model on the QW dataset to quantitatively measure fluency of query sequences. After fine-tuning, the well-formedness model achieves 42.32\% for 6-way classification on the test set. 
The QW model predicts 0.0 and 1.0 with 72\% and 95\% accuracy respectively. Intermediate scores classes are less accurate. The model puts predictions of these classes in neighbouring score categories as averaged absolute difference between predictions and labels are around 0.3. This is slightly higher than the difference between two score categories. So we believe this discrepancy is acceptable. The well-formedness study \cite{Faruqui2018IdentifyingWN} focuses on binary accuracy using 0.8 as a threshold to determine whether a question is well-formed. Using the same threshold to group the multi-class predictions, the binary classification accuracy is 79.56\%, which is better than 70.7\% reported by the best model in the original paper, and close to the accuracy of 81.6\% from a BERT-based model \cite{Chhina2020WF}. Overall, we believe this model is a decent human proxy to judge the coherence and fluency of reformulations even though there is room for improvement to reach the human upper-bound binary accuracy of 88.4\% reported in \cite{Faruqui2018IdentifyingWN}. With this well-formedness model to produce fluency scores, in the last column of Table \ref{table:qw_distribution}, we see that when using QRT5 models to reformulate all queries in the SearchQA dev set, the average well-formedness scores can improve significantly. Fluency score distributions produced by QRT5 and AQA before and after the PG method are also compared, and we observe that RL can hurt well-formedness of reformulations produced by both models. However, QRT5 can retain more fluency compared to its AQA counterpart. In addition, it is worth noting that the average score of 300k raw queries in our internal log dataset is 0.5015, compared to 0.0275 of SearchQA dev set. This shows that SearchQA is a challenging set for fluency, at least more ill-formed than real-world queries.

\begin{table}
\centering
\caption{Comparison of Predicted Well-formedness Score Distributions (proportions of queries in different score categories are shown in percentages) \& Mean Scores on SearchQA Validation Set}\label{table:qw_distribution}
\begin{tabular}{|c||r|r|r|r|r|r||c|}
\hline
Model & 0.0 & 0.2 & 0.4 & 0.6 & 0.8 & 1.0 & Mean Score\\
\hline
\hline
Original Queries & 89.03 \%  & 9.49 \% & 0.70 \% & 0.36 \% & 0.36 \% & 0.07 \% & 0.0275 \\
\hline
AQA (no RL) &  55.90 \% & 18.92 \% & 8.78 \% & 3.38 \% & 6.44 \% & 6.58 \% & 0.2106 \\
QRT5 (no RL) & 5.67 \% & 7.56 \% & 6.11 \% & \textbf{6.59 \%} & \textbf{13.29 \% } & \textbf{60.78 \%} & \textbf{0.7933} \\
\hline
AQA (RL) & 88.13 \% & 9.05 \% & 1.14 \% & 0.07 \% & 0.47 \% & 1.13 \% & 0.0381 \\
QRT5 (RL) & 70.91 \% & 18.11 \% & 2.89 \% & \textbf{2.14 \%} & \textbf{3.70 \% }& \textbf{2.26 \%} & \textbf{0.1128} \\
\hline
\end{tabular}
\end{table}


\subsection{Qualitative Comparisons}
\label{qualitative_comparison}

\begin{table}[!h]
\caption{\label{table:sqa} Comparison of Reformulations by Different Methods on SearchQA Queries}
\footnotesize
\begin{tabularx}{\columnwidth}{|C|C|C|}
\hline \textbf{Model} & \textbf{Query 1} & \textbf{Query 2} \\ \hline \hline
 0. Original Query & 
 \textit{1909 nobel prize winner failed entrance exams univ bologna , italy} &
 \textit{2000 film jackie chan old west could funnier?"}
 \\ \hline
 1. Downloaded pre-trained AQA (no RL) & 
 How many won univ bologna 's italy nobel prize won? &
 Where is the chan old west west?
 \\ \hline
 2. Downloaded pre-trained AQA (with RL) & 
 What is is 1909 nobel prize winner failed entry exams univ bologna , italy name? &
 What is is is 2000 is is chan jackie chan old west might funnier name?
 \\ \hline 
 3. Policy gradient AQA baseline (PG) & 
 What is 1909 nobel prize winner failed entrance exams univ bologna , italy name name? &
 What is 2000 film jackie chan Old west might funnier name?
 \\ \hline
 4. Self-critical training AQA (SC) & 
 Name of orange nobel prize winner exam bologna ' italy name 02 name univ bologna '''''' &
 Where is chan old west west west horse is located 2000 west west pogamerum is funnier name 2000 chan chan old west west name
 \\ \hline
 5. PG + SC + fluency metric AQA &
 Where is the name of the 1909 nobel prize win entrance exams univ bologna , italy? &
 What is the name jackie chan old west is funnier from 2000 chan old west west is chan old west is it made from the is it
 \\ \hline
 6. Fine-tuned QRT5 (no RL) & How to prove that a 1909 Nobel Prize winner failed entrance exams at the University of Bosnia and Herzegovina, Italy? &
 Can the 2000 film Jackie Chan’s Old West be funnier?
 \\ \hline
 7. QRT5 with PG & What 1909 nobel prize winner failed entrance exams univ bologna, italy? &
 Is 2000 film Jackie chan old west could funnier? \\ \hline
\end{tabularx}
\end{table}

\subsubsection{SearchQA}
\label{qualitative_comparison_sqa}
Qualitatively, when given an original query, reformulation qualities can vary for different models. For each approach, we generate reformulations with the best performing model on the validation set and decode greedily. Note that in Table \ref{table:sqa}, all reformulations can get the correct answer with an F1 score of 1. We observe that the first model without RL does not understand what the question is asking for. Models 2,3, and 4 are relatively more fluent, and model 4 does more exploration in picking the words but the reformulation gets less coherent. Model 5 does relatively well on reward acquisition as shown in Figure \ref{aqa_rewards}, but it does not help the quality of this particular query even though the fluency metric is used as an extra reward. For QRT5 models 6 and 7, we notice improvements in coherence. However, before RL, QRT5 struggles to understand the purpose of the query so it reformulates the query into a general ``how" question based on its general knowledge from pre-training on a large corpus. After RL, it asks a ``what" question with minimal expansion of the original sequence. This may be due to RL reward signals revealing that these QA queries are often asking for a certain kind of entity, which in this case, is the Nobel prize winner.

With Query 2, again, all models manage to get the answer exactly with the max F1 reward. Notice that the SearchQA dataset consists of mostly ungrammatical and very noisy queries, similar to their concatenated snippets retrieved from the Google search engine. Therefore, it is difficult for our models to generalize and reformulate by transfer learning or RL-tuning alone without any additional information, a hard task even for us humans. For instance, as another example from SearchQA, a sequence of word salad: \textit{``blue river runs black forest black sea''} is almost impossible to reformulate without any context. In Table \ref{table:sqa}, model 3 trained by us is the more fluent variant with less repetitions compared to the downloaded model 2, even though the two approaches are the same. QRT5 models are constrained by the word \textit{``funnier"}, and the reformulation asks a yes-or-no question. It is curious to see that the black-box BiDAF QA environment can still give the correct answers. Therefore, as long as certain keywords are present, this QA system can give the answer. This is also mentioned in the language analysis in \cite{Buck2018AnalyzingLL}. However, in some cases, we note that the environment can never output the correct answer no matter how we try to reformulate queries as humans. This can be stemmed from the idiosyncrasy of the particular QA environment, which is trained on the noisy original dataset. Although it may not be entirely reliable, we use this pre-trained system to compare with the previous work.

\subsubsection{Internal Dataset}
\begin{table}[!h]
\caption{\label{table:qr_compare} Comparisons between AQA and QRT5 Reformulations on Internal Queries}
\footnotesize
\centering
\begin{tabularx}{\columnwidth}{|I|C|C|C|C|}
\hline
\textbf{Original Query} & \textbf{QRT5(no RL)} & \textbf{QRT5(RL)} & \textbf{AQA(no RL)} & \textbf{AQA(RL)} \\ \hline \hline
Process for One time wire & What are the best methods for creating one time wire?
 & What is the process for a one time wire? & Process for one time wire?
 & What is one time wire process for one time wire name name…
\\ \hline 
this work object hasnt been touched in two weeks and is just sitting there & What can I do to solve this working object that hasn't been touched in two weeks and is sitting there? & What to do about a work object that hasn't been touched in two weeks and is just sitting there? & This work object has been touched in two weeks and just sitting?
 & What is this work object hasnt been touched in two weeks and just sitting have just name?
 \\ \hline
how to reinvest for an account? & How to fund reinvestments to an account?
 & How do I reinvest for an account?
 & How do you reinvest a account?
 & What is how to reinvest a account? \\ \hline
\end{tabularx}
\end{table}
We sample from the internal log dataset of queries and evaluate AQA models and QRT5 models' generalization on this out-of-domain dataset before and after RL from SearchQA. For each input query, we generate multiple reformulations with beam search and pick ones with the best qualities. We notice that when original queries are already well-formed and complete, T5 makes minimal changes. Otherwise, we get reformulations that transform and expand incomplete queries into proper questions that are aligned with the original intent semantically as shown in Table \ref{table:qr_compare}. Compared with QRT5, it is clear that out-of-sample reformulations generated by AQA reformulators are more rigid and prone to repetitions and errors before and after RL on SearchQA, which is also corroborated in Table \ref{table:sqa}. This phenomenon is addressed in \cite{Buck2018AskTR} and \cite{Buck2018AnalyzingLL}, where the reformulated language is regarded as an instance of machine-to-machine translation, and repetitions are acceptable through the lens of information retrieval. However, qualitative analysis suggests that AQA models that learn to communicate between machines are not adequate in the case when reformulations need to be fluent enough to be shown, and examined by human users in a client-facing QA setting.

\subsection{Intent Classification}

\begin{table}
\centering
\caption{Intent Classification Performance with QRT5: QRT5-IC (no RL) represents test performance using naïve reformulations with a fine-tuned QRT5 model, QRT5-IC (with RL) uses the PG method to learn from engineered IC reward signals}\label{table:intent}

\begin{tabular}{|c||c|c|c|}
\hline
Metrics & Original IC System & QRT5-IC (RL) & QRT5-IC (no RL) \\
\hline
\hline
Accuracy & 0.6010 & \textbf{0.6207} & 0.5883 \\
\hline
F1 Score &  0.6064 & \textbf{0.6267} & 0.5941 \\
\hline
\end{tabular}
\end{table}

In Table \ref{table:intent}, we find that naive reformulations hurts the performance in accuracy and F1 scores. After RL, the accuracy and F1 scores of the IC system can be improved with reformulated queries from QRT5. This means even though supervised fine-tuning of QRT5 before RL can improve the fluency of queries by a large margin as shown in Table \ref{table:qw_distribution}, naïve reformulations prehaps drift away from original intended purposes due to the discrepancy between the fine-tuning data (MQR and Quora) and OOD internal query logs. The model may emphasize on generation quality excessively during the fine-tuning stage at a cost of losing original intents. Thus, it pays to further leverage RL for the reformulator to adapt to reward signals from downstream environments like IC. This can correct the course from semantic drift, even though in the RL process, sequence fluency is impaired as the trade-off.

 \section{Related Work}
Our work is similar to the task of paraphrasing. This is to restate a given sequence to preserve the same meaning. \cite{Witteveen_2019} focus on fine-tuning GPT-2 \cite{Radford2019LanguageMA} with supervised datasets. Algorithmic n-gram metrics are used to find the best paraphrase. The models demonstrate the ability to generalize on OOD text. Another common approach is to leverage machine translation. \cite{Guo2019ZeroShotPG}'s work uses multilingual translation to pivot for zero-shot paraphrases. Human evaluators are employed in this process to evaluate fluency. \cite{Roy2019UnsupervisedPW} uses a VQ-VAE \cite{Oord2017NeuralDR} to compare a monolingual paraphrasing method with unsupervised and supervised translation approaches, among other generative techniques \cite{Yang2019AnEG,Gupta2018ADG}. Using Deep RL to paraphrase has been attempted. With reward shaping and policy gradient in \cite{li-etal-2018-paraphrase}, a generator paraphrases by learning rewards from an trained evaluator that can judge whether two sentences are paraphrases. Transitioning from an unsupervised VAE model to RL is possible on non-parallel data with reward engineering based on characteristics of good paraphrases \cite{Siddique_2020}. However, these studies merely focus on generation quality rather than gaining performance on any downstream systems. Our RL framework aims to retain reformulation quality as well as performance in downstream applications. This is most related to \cite{Buck2018AskTR}'s AQA approach by leveraging policy-based RL methods to generate question reformulations with a GNMT reformulator \cite{Wu2016GooglesNM} and a CNN \cite{Krizhevsky2012ImageNetCW} selector. The reformulations are treated as inputs to a BiDAF \cite{Seo2017BidirectionalAF} QA system that generates character-level F1 rewards. Rather than a translation-based model, our approach fine-tunes T5 \cite{Raffel2019ExploringTL} to leverage the flexibility in this framework and the linguistic prior encoded by the text-to-text model from pre-training. \cite{Lin2020ConversationalQR} has shown T5's good performance on reformulating questions along with context within a conversational history. While in our work, only original queries are considered as input to the black-box QA environment during RL for fair comparison. Identifying well-formed questions \cite{Faruqui2018IdentifyingWN} by training binary classification models has been studied using BERT \cite{Chhina2020WF} and transfer learning with pre-trained models \cite{Syed2019InductiveTL}. We investigate more fine-grained 6-way classification using a fined-tuned T5 for regression, leveraged as a proxy to evaluate fluency of reformulations.


\section{Conclusions}
We show that QRT5 can optimize the RL objective to reformulate queries, adapting to reward signals sourced from black-box systems like question answering and intent classification. RL variants based on policy-gradient and self-critical training can achieve better downstream performance compared to other alternatives. We find during RL, reformulators struggle to maintain the ability to generate fluent queries compared to before RL. Transfer learning under the text-to-text framework has proven critical for reformulation models to retain fluency. It provides flexibility to fine-tune on paraphrasing and denoising, as well as creating a prediction model to evaluate fluency on reformulations, driven by human evaluations. The fine-tuned QRT5 model is capable of generating reformulations with quality for out-of-sample queries before RL. This provides a more robust starting point than the previous approach for later RL tuning. After RL in a specific black-box environment such as QA, QRT5 demonstrates its ability to maintain fluency qualitatively and quantitatively while acquiring rewards from the downstream task. Finally, as text-to-text is more flexible in nature, swapping out the QA systems, systematically studying generalization on other out-of-domain datasets, and adding conversational context to produce more informed reformulations are promising future directions.

\bibliography{citation}
\bibliographystyle{splncs04}





\end{document}